\documentclass[10pt,twocolumn,letterpaper]{article}

\usepackage{iccv}              

\usepackage{algorithm}
\usepackage{algorithmic}
\usepackage{multirow} 
\usepackage{booktabs} 
\usepackage{amsmath} 
\usepackage{amssymb} 
\usepackage{bbm}
\usepackage{newtxmath}

\definecolor{iccvblue}{rgb}{0.21,0.49,0.74}
\usepackage[pagebackref,breaklinks,colorlinks,allcolors=iccvblue]{hyperref}

\def\paperID{8358} 
\def\confName{ICCV}
\def\confYear{2025}
\usepackage{color}
\definecolor{green}{rgb}{0, 0.5, 0}
\definecolor{orange}{rgb}{0.8, 0.6, 0.2}
\definecolor{orange2}{rgb}{1.0, 0.6, 0.2}
\definecolor{red}{rgb}{1.0, 0.0, 0.0}
\definecolor{teal}{rgb}{0.0, 0.4, 0.4}
\definecolor{purple}{rgb}{0.65,0,0.65}
\definecolor{saffron}{rgb}{0.95,0.75,0.2}
\definecolor{turquoise}{rgb}{0.0,0.5,0.5}
\definecolor{black}{rgb}{0.0, 0.0, 0.0}
\definecolor{gray}{rgb}{0.5, 0.5, 0.5}

\newcommand{\modify}[1]{{\color{black}#1}}

\title{MoFRR: Mixture of Diffusion Models for Face Retouching Restoration}

\author{Jiaxin Liu\textsuperscript{1}, Qichao Ying\textsuperscript{1}, Zhenxing Qian\textsuperscript{1}, Sheng Li\textsuperscript{1$\dagger$}, Runqi Zhang\textsuperscript{1}, Jian Liu\textsuperscript{2}, Xinpeng Zhang\textsuperscript{1}\\
\textsuperscript{1} Fudan University
\textsuperscript{2}Ant Group\\
}

\begin{document}
\maketitle
\begin{abstract}
The widespread use of face retouching on social media platforms raises concerns about the authenticity of face images. While existing methods focus on detecting face retouching, how to accurately recover the original faces from the retouched ones has yet to be answered. This paper introduces Face Retouching Restoration (FRR), a novel computer vision task aimed at restoring original faces from their retouched counterparts. FRR differs from traditional image restoration tasks by addressing the complex retouching operations with various types and degrees, which focuses more on the restoration of the low-frequency information of the faces. To tackle this challenge, we propose MoFRR, Mixture of Diffusion Models for FRR. Inspired by DeepSeek's expert isolation strategy, the MoFRR uses sparse activation of specialized experts handling distinct retouching types and the engagement of a shared expert dealing with universal retouching traces. Each specialized expert follows a dual-branch structure with a DDIM-based low-frequency branch guided by an Iterative Distortion Evaluation Module (IDEM) and a Cross-Attention-based High-Frequency branch (HFCAM) for detail refinement. Extensive experiments on a newly constructed face retouching dataset, RetouchingFFHQ++, demonstrate the effectiveness of MoFRR for FRR.
\end{abstract}    
\section{Introduction}
\label{sec:intro}

\newcommand\blfootnote[1]{%
\begingroup
\renewcommand\thefootnote{}\footnote{#1}%
\addtocounter{footnote}{-1}%
\endgroup
}
\blfootnote{$\dagger$Corresponding author: Sheng Li (lisheng@fudan.edu.cn)}

\begin{figure}[tb]  
    \centering
    \includegraphics[width=1\linewidth]{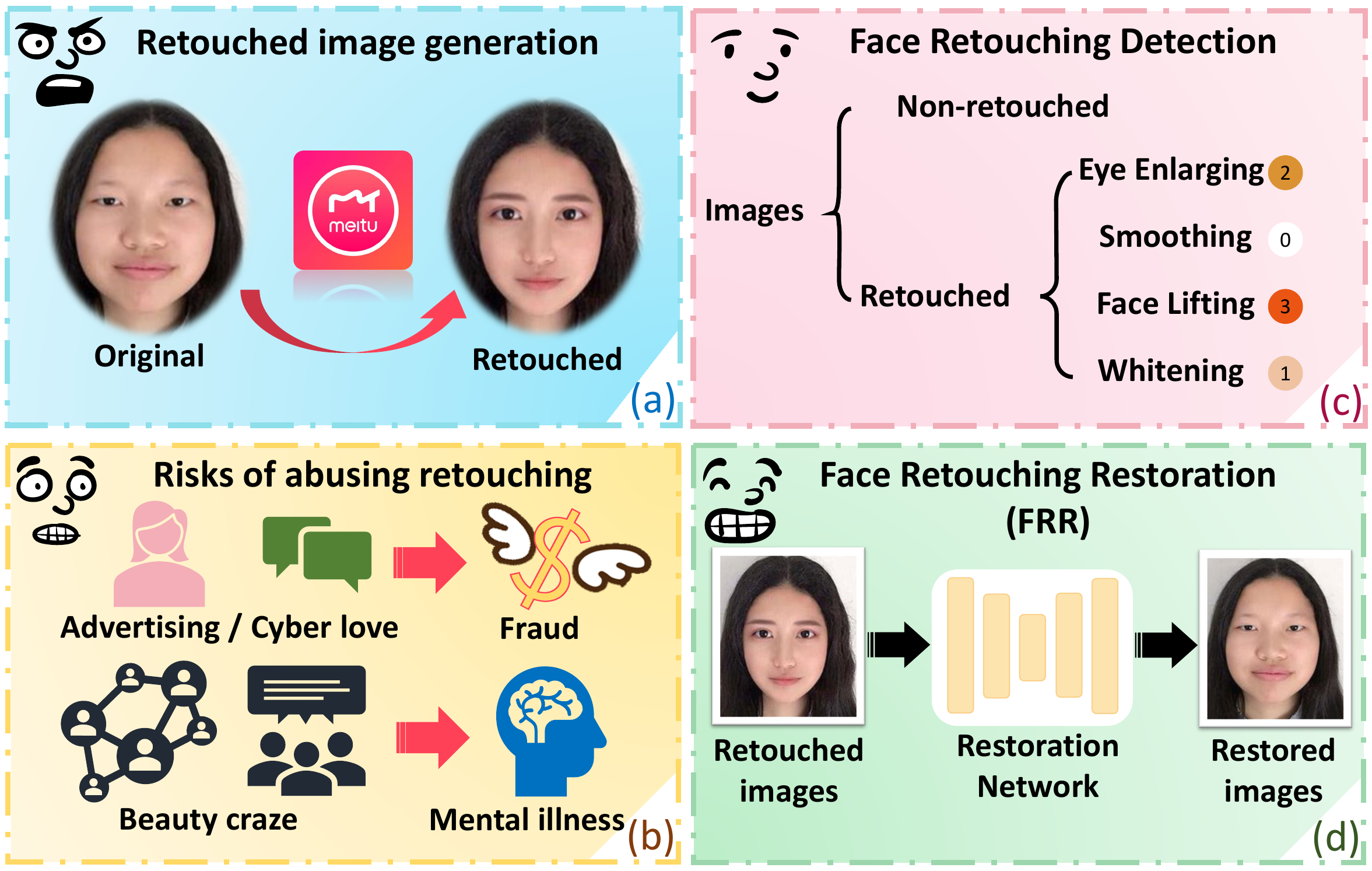}
    \caption{Application scenario of the proposed scheme (MoFRR). (a) People use face retouching in various applications, (b) face retouching poses risks such as fraud, societal security, and cultural psychological issues, (c) existing work for face retouching detection, (d) our proposed MoFRR method recovers the original image from the retouched version, thereby offering an additional layer of protection against face retouching fraud.}
    \label{fig:head}
\end{figure}

Retouched face images, produced through various techniques such as face lifting, eye enlargement, whitening, and smoothing, are pervasive on social platforms. Most users acquire these images through straightforward and convenient methods to improve their appearance. Despite amusement and fun, the abuse of face retouching causes a lot of serious problems, including aesthetic degradation, commercial deception~\cite{ateq2024association} and identity fraud~\cite{bharati2016detecting}.
To address these problems, many retouching detection schemes and regulatory countermeasures have been proposed. For instance, Norway~\cite{c:norway} has enacted strict disclosure requirements for edited promotional content. US and Israel also
have labeling requirements for retouched faces~\cite{USlaw,Israellaw}.
Researchers have developed dedicated datasets and neural networks to auto-detect the presence of retouching within a digital image~\cite{ying2023retouchingffhq,VMUYMU,MIW}. 
In application, once exaggerated retouching is detected, the corresponding images can be automatically removed, or the livestreams closed.

In the literature, researchers have focused mainly on the accurate detection of face retouching ~\cite{rathgeb2020prnu,jain2020detecting, ying2023retouchingffhq}. It has yet to be answered regarding how we could near-faithfully recover the original face image from the retouched one, which is important to trace back to the real identity of the severely retouched faces. 

In this paper, we consider a new computer vision task, Face Retouching Restoration (FRR), to restore the original faces from the retouched ones, which serves as a further step for fighting against the misuse of face retouching. In literature, researchers have dedicated their efforts to dealing with similar tasks, including makeup removal~\cite{2016facebehindmakeup} and image restoration (IR)~\cite{khan2023pasd,lin2024diffbir, OSEDiff}. The former transfers the style of a reference face to a makeup face for face restoration. The latter tries to restore high-quality textural details from low-quality images. These schemes work well in recovering textural features. However, in face retouching, it may not be sufficient to change only the textural features of the face image. Usually, the face structure has to be modified in order to have distinct differences in appearance. Consequently, noticeable structural changes can be made during face retouching, such as enlarged eyes, more prominent noses, and thinner face shapes.  
Thus, directly applying existing makeup removal or image restoration schemes may not be appropriate for FRR.

We believe that a good and appropriate FRR scheme should focus more on the restoration of low-frequency information of the faces, which significantly differs from the existing approaches for the IR and makeup removal tasks. FRR is a challenging task as the faces might be operated in a complex setting of face retouching with various types and degrees, including whitening, smoothing, eye enlargement, and face lifting. The logic among different touching types varies and the goals are also different. Thus, it might not be appropriate to perform the FRR using a single model. We also notice that the impact of different face retouching operations is independent in the face images.

Inspired by the Mixture of Experts (MoE)~\cite{jordan1994MoE,cai2024MoEsurvey}, we propose MoFRR that restores the face images in a divide-and-conquer manner using an MoE framework with specialized and shared experts. A specialized expert is trained to explicitly restore the faces for a certain type of face retouching. The design of the shared expert is inspired by the recent success of DeepSeek~\cite{liu2024deepseek}, which aims to deal with universal retouching traces in different types of face retouching. We propose a novel wavelet DDIM~\cite{song2020DDIM} model, denoted as WaveFRR, for the design of specialized experts. Each WaveFRR contains a dual-branch structure including a DDIM-based low-frequency branch and a cross-attention-based high-frequency branch. 
The former contains an Iterative Distortion Evaluation Module (IDEM) that restores the wavelet low-frequency sub-band of the faces. The latter contains a High-Frequency Cross-Attention Module (HFCAM) that recovers the wavelet high-frequency sub-bands. For the shared expert, we adopt the ordinary DDIM architecture to make it general for various retouching conditions. 

For training and evaluation, we newly construct a dataset named RetouchingFFHQ++, which contains over a million retouched face images from four commercial face retouching APIs. We have conducted comprehensive experiments including intra-API and cross-API tests, both demonstrate the effectiveness of our MoFRR for face retouching restoration. To summarize, our contributions are as follows.

\begin{itemize}
    \item We are the first to consider the task of face retouching restoration, and we propose MoFRR, an MoE framework with both specialized and shared experts, to address this challenging problem.
    \item We propose a wavelet-based DDIM model for face restoration from a specific retouching type, where an IDEM module and an HFCAM moule are designed to recover the low-frequency and high-frequency sub-bands of the face, respectively.
    \item We extend the large-scale face retouching dataset, RetouchingFFHQ~\cite{ying2023retouchingffhq}, into RetouchingFFHQ++ to make it more suitable for training and evaluation of FRR schemes. 
\end{itemize}
\section{Related Works}
\label{sec:related_works}


\subsection{Makeup Transfer and Removal}
Over the past year, makeup removal~\cite{2016facebehindmakeup,chen2017makeup} has garnered significant attention in research, often explored in conjunction with makeup transfer~\cite{gu2019ladn}. 
PairedCycleGAN~\cite{chang2018pairedcyclegan} utilizes an asymmetric framework that includes an additional sub-network specifically for makeup removal. 
PSGAN++~\cite{liu2021psgan++} employs a makeup distillation network and an identity extraction network to facilitate makeup transfer and removal. 
SSAT~\cite{sun2022ssat} is proposed for simultaneous makeup transfer and removal, utilizing a Symmetric Semantic Corresponding Feature Transfer module for accurate semantic alignment.
Recently, CSD-MT~\cite{sun2024CSD-MT} introduces an unsupervised framework to achieve robust performance by adaptively combining high-frequency content details and low-frequency style features.
Nevertheless, face retouching differs from making-up in that the former often involves changes to the facial structure while the latter does not.

\subsection{Image Restoration}
Many methods are proposed to utilize facial component dictionaries~\cite{li2020dfdnet} or codebooks~\cite{gu2022vqfr,wang2023restoreformer++} derived from high-quality images to guide the restoration process. 
DR2~\cite{wang2023dr2} uses a DDPM to transform degraded images into coarse, degradation-invariant predictions, which are then enhanced to high-quality images. 
Restormer~\cite{zamir2022restormer} uses an efficient Transformer to achieve superior performance on several image restoration tasks. 
ResDiff~\cite{shang2024resdiff} proposes a DDPM to predict residuals, using frequency-domain loss and guided diffusion to improve generation speed and sample quality.
However, image restoration primarily focuses on recovering high-frequency texture details for image quality improvement, while FRR often aims to restore a face image more semantically.

\begin{figure}[!t]
    \centering
    \includegraphics[width=1\linewidth]{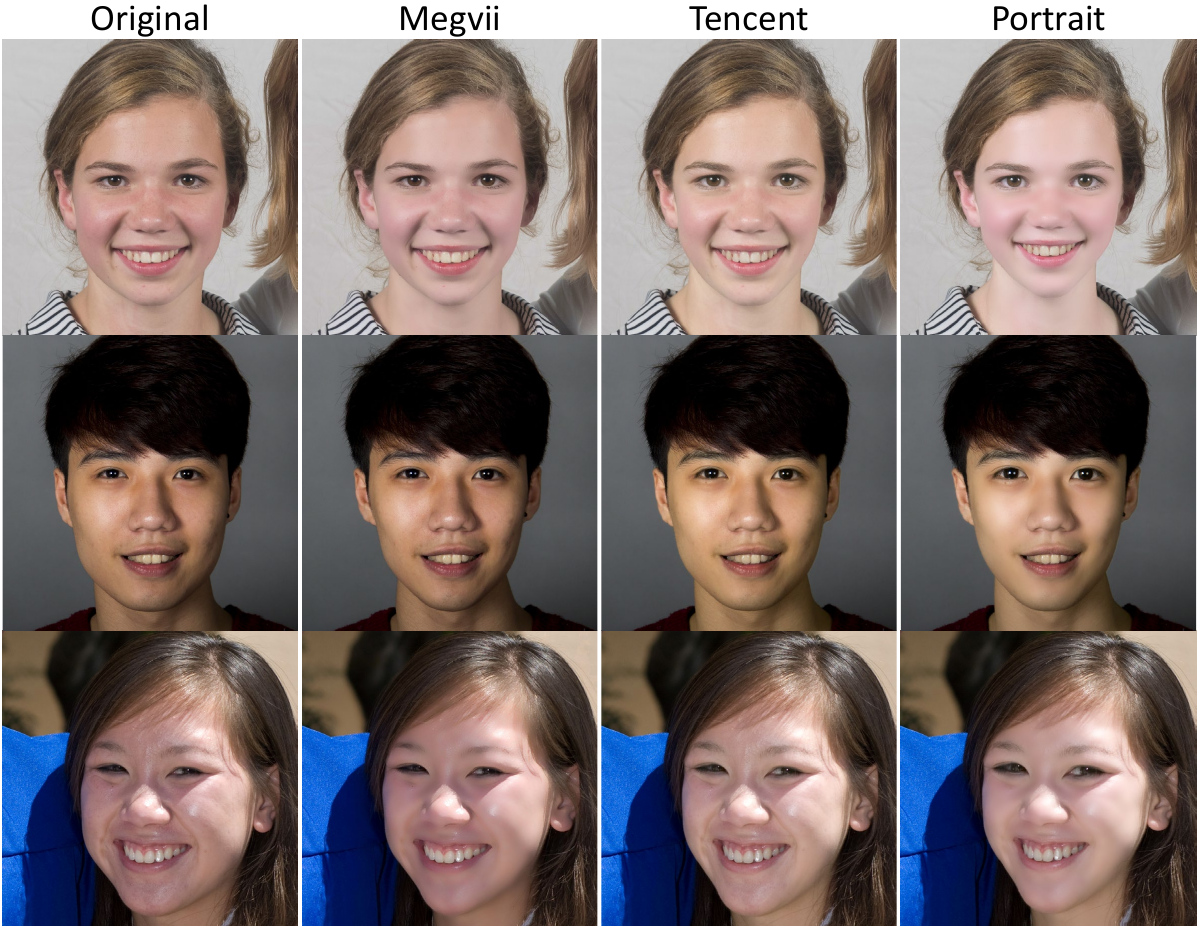}
    \caption{Examples from the RetouchingFFHQ++ dataset, where original faces are retouched by different retouching APIs. In each row, the applied retouching types and degrees are kept same.}
    \label{fig:api}
\end{figure} 
\begin{table}[!t] 
\centering
\footnotesize
\setlength{\tabcolsep}{4pt} 
\begin{tabular}{c cccc c} 
\toprule
& Portrait & Megvii & Tencent & Alibaba & Total\\ 
\midrule
Ori & - & - & - & - & 57910\\
\midrule
Single & 108267 & 66692 & 200100 & 29483 & 404542\\
\midrule
Multi & 206655 & 232641 & 173590 & - & 612886\\ 
\midrule
Total & 314922 & 299333 & 373690 & 29483 & 1075338\\
\bottomrule
\end{tabular}
\caption{Summary of RetouchingFFHQ++. The ``Multi" column includes a mixture of 2+ face retouching operations.}
\label{tab:dataset num}
\end{table}

\section{Method}
\label{sec:method}


\subsection{FRR: Goals, Evaluations and Our Dataset}
\modify{
\noindent\textbf{Goals and Evaluations}. 
The goal of FRR is defined to blindly, i.e., with no other reference of template, reconstruct the original face images given the retouched ones via computer vision algorithms or deep learning methods.
Our evaluation framework prioritizes forensic admissibility over perceptual quality, explicitly discouraging methods that synthesize plausible but inauthentic facial features. We adopt a dual-metric approach: pixel-level fidelity is quantified through standard PSNR and SSIM~\cite{wang2004ssim} to measure spatial and structural reconstruction accuracy, while biometric veracity is assessed via feature-space cosine similarity using state-of-the-art face recognition models like AdaFace~\cite{kim2022adaface} and ArcFace~\cite{deng2019arcface}, etc.
This dual evaluation mechanism ensures FRR methods meet both visual reconstruction benchmarks and evidentiary standards required for judicial applications, where synthetic enhancements could render biometric evidence legally inadmissible.
}

\noindent\textbf{The RetouchingFFHQ++ Dataset.}
In order to train networks for FRR, we prepare our first dedicated dataset for FRR based on RetouchingFFHQ~\cite{ying2023retouchingffhq}, a recently proposed large-scale face retouching detection dataset that includes high-quality, fine-grained annotated retouched face images from commercial APIs such as Tencent, Megvii, and Alibaba.
The original face images are from the famous FFHQ dataset~\cite{FFHQ}, and the retouched subsets includes thousands of retouched images, covering whitening, smoothing, face lifting, eye enlarging, or combined operations of the above.
\modify{
But the dataset is originally proposed for face retouching detection, so in order to better cater to our FRR task, we extend this dataset as RetouchingFFHQ++ in the following two aspects.


First, we expand the dataset using the widely-adopted PortraitPro 24 API~\cite{anthropics_portraitpro}, increasing the total number of images to over a million, as shown in \cref{tab:dataset num}.
We show examples of the dataset in \cref{fig:api}, and
we include more details and statistics of the dataset in the supplement. 
\modify{
From the exampled images, we see that retouched images by the PortraitPro API can have noticeable visual differences compared to the existing images from other APIs, thus further diversifying the dataset.
}

Second, we propose a more reasonable criteria on the degree definition of each retouching.
While RetouchingFFHQ directly defines the degrees according to the argument passed on generating the images, it could less reflect the actual degree due to API-level algorithmic discrepancy.
To mitigate this, we redefine the degree of modification of each operation within images via statistically analyzing the PSNR distribution of all retouched face images.
We re-categorize them into five groups with proportions of $15\%, 25\%, 25\%, 25\%$, and $10\%$, based on ascending PSNR values, and labeled from 5 to 1 accordingly.
}

\begin{figure*}[!t] 
    \centering
    \includegraphics[width=1\linewidth]{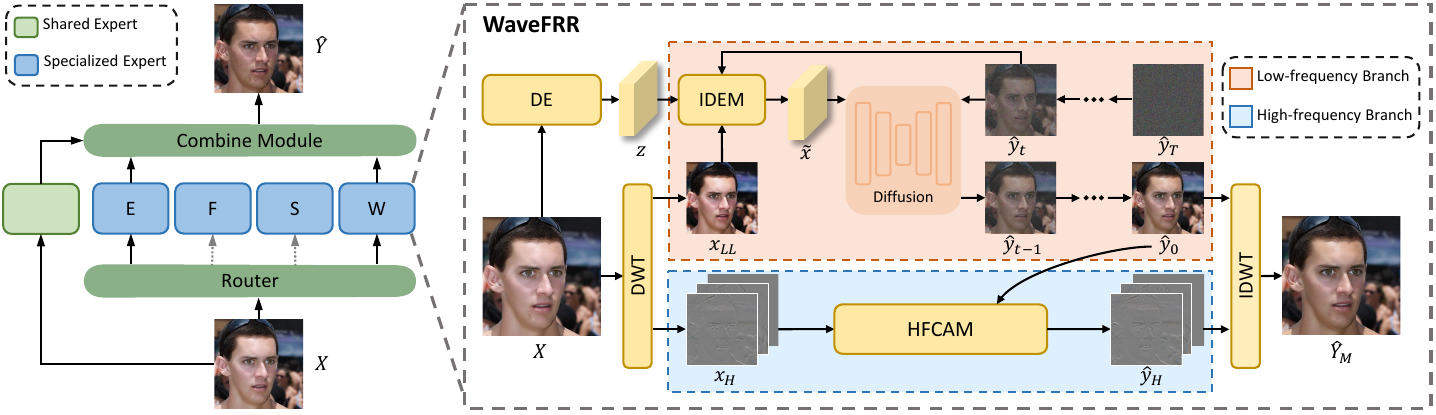}
    \caption{Left: Overview of MoFRR, where input image is processed by the router and selectively sent into the specialized experts for different retouching types. A shared expert is continuously engaged inspired by Deepseek, and the result images from the experts are jointly sent to a lightweight Combine Module to produce the final restored image. Right: the WaveFRR design, where the image is decomposed using DWT, and the DDIM process works on low-frequency subband restoration guided by the transformed degree estimation result (by IDEM). Then HFCAM module restores the high-frequency sub-bands via cross-attending the restored lower sub-bands.}    
\label{fig:expert}
\end{figure*}

\modify{\subsection{FRR Model with Multi Experts}}

The overall pipeline of our proposed MoFRR, as depicted in \cref{fig:expert}, is a Mixture of Diffusion Models for FRR.
We explicitly decompose the task into several steps. 
First, given a targeted image, we first predict which types of face retouching operations are performed. 
Second, we employ several expert networks, where apart from a reserved shared expert as inspired by DeepSeek, the rest of experts (denoted as specialized experts) are respectively specialized in removing one certain type of face retouching in the given image. As a result, the experts provide several versions of intermediate recovered images. 
Third, with the intermediate images from all the activated experts, we merge these images with the original image through the Combine Module to obtain the final restored image. 

Concretely, we train a \modify{router} to identify which face retouching operations have been applied to the images and to activate the following expert networks.
We implement it by the ResNet-MAM~\cite{ying2023retouchingffhq}, which is adapted for multi-label classification and outputs an $N$-dimensional binary label vector. Here in our paper, we focus on the most typical types of face retouching operations, which are \textit{whitening, smoothing, face lifting}, and \textit{eye enlarging}, so $N=4$.
\begin{equation}
[b_w,b_s,b_f,b_e] = \mathbbm{1}({\text{Router}(X)>\text{th}}),
\label{eq:my_equation}
\end{equation}
where $X\in\mathbb{R}^{H\times~W\times~C}$ represents the input retouched face image.
$b_M$ for $M\in\{w,s,f,e\}$ denotes the prediction result of whether the input image contains the corresponding face retouching operation $M$. $\mathbbm{1}({\cdot>\text{th}})$ represents if the prediction exceeds a given threshold (here 0.5 by default).

Next, four ($N$) independent experts, named WaveFRR models, process the image once activated by the router. 
The design of a WaveFRR model is illustrated in \cref{fig:expert}. 
Each expert is specialized in estimating and restoring a particular face retouching operation. 
It begins with predicting the degree of the specific face retouching operation, which is trained ahead and guides the subsequent restoration process.
Then,
considering that face retouching primarily emphasizes content modifications, such as cheeks and eyes, rather than global high-frequency details, we first perform the discrete wavelet transformation (DWT) that decomposes the input image into lower and higher frequencies.
For lower frequencies, we propose sampling a retouching-free version via conditional DDIM~\cite{ho2020DDPM,song2020DDIM},
guided by the predicted retouching degree and the original lower frequencies.
For higher frequencies, we modify them by cross-attending~\cite{ViT} the generated lower frequencies for more refined retouching removal. 
The modified frequencies are then combined via Inverse Discrete Wavelet Transform (IDWT) to produce the restored image $\hat{Y}_M$, which removes only the specific face retouching operation $M$. 

\modify{For the shared expert, we adopt a baseline DDIM architecture without wavelet decomposition, where we hope the architectural divergence from WaveFRR explicitly could promote complementary functional specialization. 
While WaveFRR experts focus on frequency-aware sub-band recovery, the shared expert captures global retouching patterns across spectral domains, denoted as $\hat{Y}_{se}$.} 

Finally, we gather all restored outputs of the activated experts and design a lightweight Combine Module to give the final retouching-free image $\hat{Y}$. 
The Combine Module employs a UNet~\cite{ronneberger2015UNet}, which leverages all images and the retouched image to generate results while preserving fidelity to the input image, as follows:
\begin{equation}
\hat{Y} = \text{Combine\_Module}([X, \hat{Y}_{se}, \hat{Y}_{M_0},\hat{Y}_{M_1}, ...]).
\label{eq:my_equation}
\end{equation}


\subsection{WaveFRR Model}

\noindent\textbf{Degree Estimator and DWT.}
Each expert begins with estimating the degree of its specialized face retouching operation.
Afterward, we train a ResNet50~\cite{he2016resnet} as Degree Estimators, using our newly defined degree labels as supervision. 
It predicts the global retouching degree $z=\text{DE}_M(X)$ for the face retouching operation $M$. 
Besides the estimator, we apply the DWT to decompose it into low-frequency and high-frequency sub-bands, denoted as $\{x_\emph{LL},x_H\}$,
where $x_\emph{LL}\in \mathbb{R}^{\frac{H}{2}\times\frac{W}{2}\times C}$ and $x_{H}\in \mathbb{R}^{\frac{H}{2}\times\frac{W}{2}\times 3C}$ represent one low-frequency sub-band, i.e., LL, and three high-frequency sub-bands, i.e., HH, LH and HL. 

\noindent\textbf{Low-Frequency Branch.}
The LL sub-band of DWT contains major information of ${X}$. 
We propose to sample a retouching-free LL sub-band of ${X}$ via conditional DDIM, guided by the predicted retouching degree and the original LL sub-band. 
For denoised image $\hat{y}_t$ at timestamp $t$, our IDEM module utilizes the multi-scale channel attention network~\cite{dai2021mscam}, denoted as $\text{MCA}(\cdot)$, which produces the pixel-wise condition $\tilde{x}\in\mathbb{R}^{\frac{H}{2}\times\frac{W}{2}\times 2C}$ given $\hat{y}_t$, $x_\emph{LL}$ and $z$ as:
\begin{equation}
F = \text{MCA}(z+x_\emph{LL})\otimes x_\emph{LL} + (1-\text{MCA}(z+x_\emph{LL}))\otimes z,
\label{eq:my_equation}
\end{equation}
\begin{equation}
\hat{R} = \text{MCA}(F+\hat{y}_t)\otimes \hat{y}_t + (1-\text{MCA}(F+\hat{y}_t))\otimes F,
\label{eq:my_equation}
\end{equation}
\begin{equation}
\tilde{x} = \text{IDEM}(\hat{y}_t, x_\emph{LL}, z)= \text{Concat}(x_\emph{LL},\hat{R}),
\end{equation}
where  $\hat{R} \in \mathbb{R}^{\frac{H}{2}\times\frac{W}{2}\times C}$ is the pixel-wise distortion map.

The DDPM~\cite{ho2020DDPM} uses a Markov chain to restore the image step by step from Gaussian noise. 
The forward diffusion process progressively transforms the ground truth $y_0$ into noise data in $T$ steps with a variance schedule $\{\beta_1,\beta_2,\ldots,\beta_T\}$, which can be formulated as:
\begin{equation}
q(y_t\mid y_{t-1})=\mathcal{N}(y_t;\sqrt{1-\beta_t}y_{t-1},\beta_t \mathbf{I}),
\label{eq:my_equation}
\end{equation}
where $\mathcal{N}$ denotes the Gaussian distribution.
The denoising process gradually removes noise from a randomly sampled Gaussian noise $\hat{y}_T\sim\mathcal{N}(0,\mathbf{I})$. We use conditional denoising process with the condition $\tilde{x}$, i.e,
\begin{equation}
p_{\theta}(\hat{y}_{0:T}\mid\tilde{x})=p(\hat{y}_T)\prod_{t=1}^T p_\theta(\hat{y}_{t-1}\mid\hat{y}_t,\tilde{x}).
\label{eq:my_equation}
\end{equation}

The sampled lower-frequent subband at the last timestamp is denoted as $\hat{y}_0$, which is regarded as the retouching-free lower frequency of $X$. \modify{To accelerate inference, we take DDIM~\cite{song2020DDIM} as the sampling scheme.}


\noindent\textbf{High-Frequency Branch and IDWT.}
We further introduce the HFCAM module to adjust the high-frequency sub-bands for refined retouching removal.
The famous visual patch-wise Cross-Attention~\cite{ViT} mechanism is efficient in providing large receptive field only with several layers' effort. 
Thus, we propose to patchify $\hat{y}_0$, $x_H$, and apply a lightweight one-layer CA and a three-layered convolution network to modify the higher-frequency bands.

\begin{equation}
\hat{y}_H = x_H+\text{Conv}(\text{CA}(\hat{y}_0,x_H)).
\label{eq:my_equation}
\end{equation}

Finally, we combine the restored frequencies via the inverse wavelet transformation, i.e., $
\hat{Y}_M = \text{IDWT}(\hat{y}_0,\hat{y}_H)$.

\subsection{Objective Functions and Training Details}

To train the proposed method, 
we first separately train the router and the Degree Estimators ahead of the rest modules for restoration.
Next, these models are fixed and we train the low- and high-frequency branches within WaveFRR. 
Once the experts are prepared, we train the Combine Module that gets the intermediate images and produces the final result.
Note that while we train MoFRR based on hybrid face images that have come through 0-4 face retouching operations, we only train WaveFRR models using single-operated images since they only remove their selected type, \modify{and train the shared expert on subset B for capturing global retouching patterns.}

The objective functions of the proposed method include three categories, namely, frequency loss $\mathcal{L}_\emph{freq}$, spacial loss $\mathcal{L}_\emph{space}$, and classification loss $\mathcal{L}_\emph{class}$.
The total loss for the WaveFRR is given by
$
\mathcal{L} = \mathcal{L}_\emph{freq}+\mathcal{L}_\emph{space}+\mathcal{L}_\emph{class}.
$

The frequency loss $\mathcal{L}_\emph{freq}$ includes three parts, namely, IDEM loss $\mathcal{L}_\emph{IDEM}$, diffusion loss $\mathcal{L}_\emph{simple}$, and high-frequency loss $\mathcal{L}_\emph{high}$.
, i.e.,
$
\mathcal{L}_\emph{freq} = \mathcal{L}_\emph{IDEM}+\mathcal{L}_\emph{simple}+\mathcal{L}_\emph{high}.
$
The IDEM loss uses the $\ell_2$ loss to quantify the difference between the pixel-wise distortion and the true residual of the low-frequency sub-band, which is given by
$
\mathcal{L}_\emph{IDEM} = \parallel\hat{R}-(y_\emph{LL}-x_\emph{LL})\parallel^2,
$
where $y_\emph{LL}$ is the LL sub-band of the ground truth image.
The training objective for diffusion models is
$
\mathcal{L}_\emph{simple} = \mathbb{E}_{y_0,t,\epsilon_t\sim \mathcal{N}(0,\mathbf{I})}\left[\parallel\epsilon_t-\epsilon_\theta(y_t,\tilde{x},t)\parallel^2\right].
$
We utilize the $\ell_2$ loss and Total Variation (TV)~\cite{chan2011TV} loss to optimize the HFCAM, defined as:
$
\mathcal{L}_\emph{high} = \lambda_1\parallel \hat{y}_H-y_H\parallel^2+\lambda_2 \text{TV}(\hat{y}_H),
$
where $\lambda_1$ and $\lambda_2$ are set as 0.1 and 0.01, respectively.

The spatial loss $\mathcal{L}_\emph{space}$ is to maximize the PSNR between restored and original image. 
We use a hybrid $\ell_1$ \& SSIM~\cite{wang2004ssim} loss, 
defined as $\mathcal{L}_\emph{hyb}(\cdot,\hat{\cdot})=\parallel{\cdot}-\hat{\cdot}\parallel_1+(1-\text{SSIM}({\cdot},\hat{\cdot}))$, where the symbol $\cdot$ denotes the ground-truth value of the symbol $\hat{\cdot}$,
to minimize the difference between the restored face image and the ground truth image for both WaveFRR and Combine Module, i.e.,
$
\mathcal{L}_\emph{space} = \mathcal{L}_\emph{hyb}(Y,\hat{Y}) + \sum_M{\mathcal{L}_\emph{hyb}(Y,\hat{Y}_M)}.
$

As for the router and Degree Estimator, we respectively train them based on cross-entropy loss $\mathcal{L}_\emph{class}$, \modify{i.e., 
$
\mathcal{L}_\emph{class} = -\sum_{i=1}^{5} c_i \log(\hat{c}_i)
$, where $c$ and $\hat{c}$ respectively denote the true label and the predicted probablility for class $i$.
}
\section{Experiments}
\label{sec:experiment}


\begin{figure*}[!t]
    \centering
    \includegraphics[width=1\linewidth]{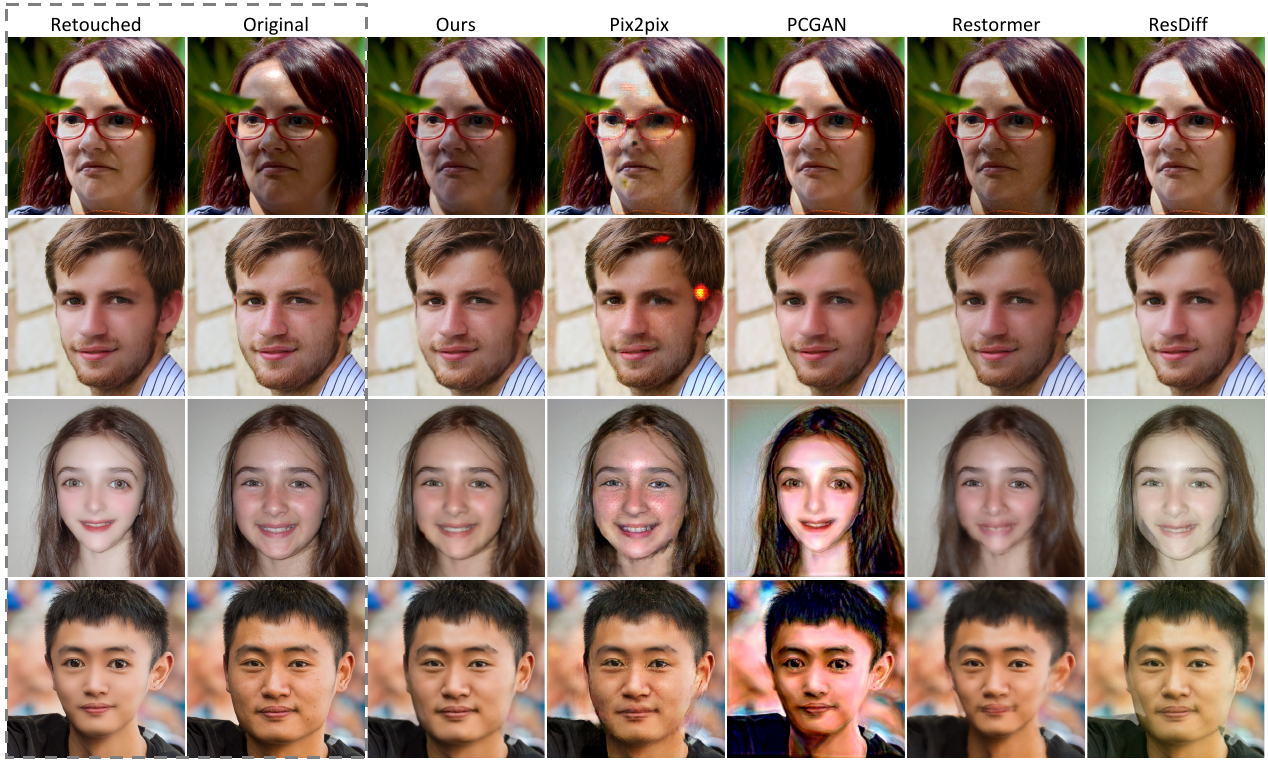} 
    \caption{Qualitative comparison of the restored faces using different methods. The images are retouched through multiple operations, and our method achieves higher restoration quality with fewer artifacts. Zoom in for best view.}
    \label{fig:comparison}
\end{figure*} 

\begin{figure}
    \centering
    \includegraphics[width=1\linewidth]{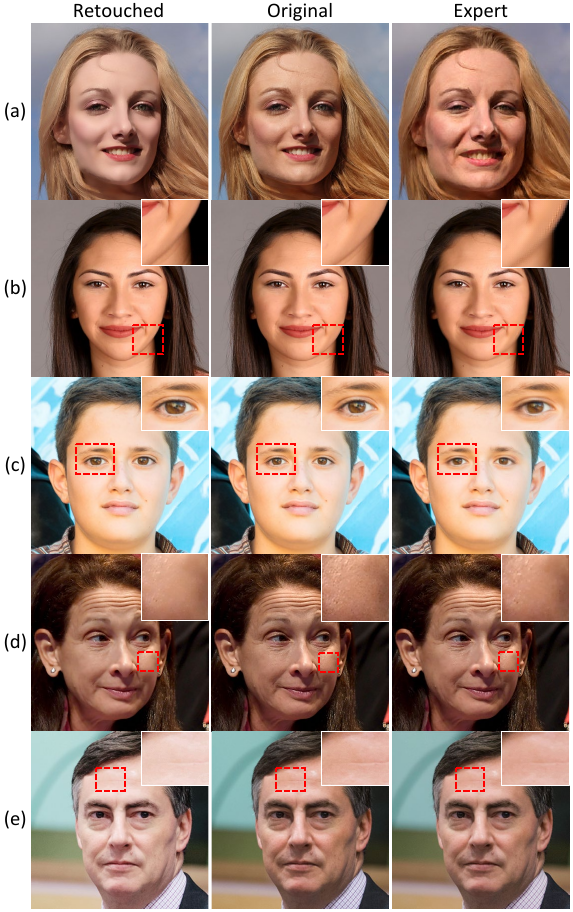}
    \caption{Visualization of images generated by five specific experts. (a) shared expert, (b) face lifting expert, (c) eye enlarging expert, (d) smoothing expert, and (e) whitening expert.} 
    \label{fig:single}
\end{figure}

\subsection{Experimental Setups}
\noindent\textbf{Implementation Details}
We use Adam~\cite{Kingma2015AdamAM} as the optimizer with a learning rate of $1\times10^{-4}$. The WaveFRR model is trained till convergence for about $2\times10^5$ iterations on eight NVIDIA A100 GPUs with a default batch size of 16, and the overall pipeline is trained for 50 epoch to converge. 
Input and output images have a resolution of $1024\times1024$. 
More details can be found in the supplement.

\noindent\textbf{\modify{Test Cases and Baseline Preparations}.}
We employ two test cases, namely, intra-API tests, and cross-API tests.
For intra-API tests, we train and test the models with a mixture of data from all APIs. 
For cross-API tests,
we train our MoFRR and all comparing models on a mixture of subsets except Portrait and test on the Portrait subset, additionally validating the generalizability of models.
In both test cases, the according dataset material is divided evenly into 8:1:1 as training/evaluation/test set.

There are no FRR methods available in the literature. 
We select a list of representative work from several relevant fields as baselines, including reference-free makeup removal method PairedCycleGAN~\cite{chang2018pairedcyclegan}  (PCGAN for short), as well as image restoration methods containing Pix2pix~\cite{isola2017pix2pix}, Restormer~\cite{zamir2022restormer}, DR2~\cite{wang2023dr2}, and ResDiff~\cite{shang2024resdiff}. The comparison methods cover a variety of mainstream architectures, including GAN, Transformer, and diffusion models.
\modify{For fair comparison,} we retrain them on our RetouchingFFHQ++ dataset for fair comparisons.




\subsection{Intra-API Performance}
\label{sec:comparison}

\begin{figure}
    \centering
    \includegraphics[width=1\linewidth]{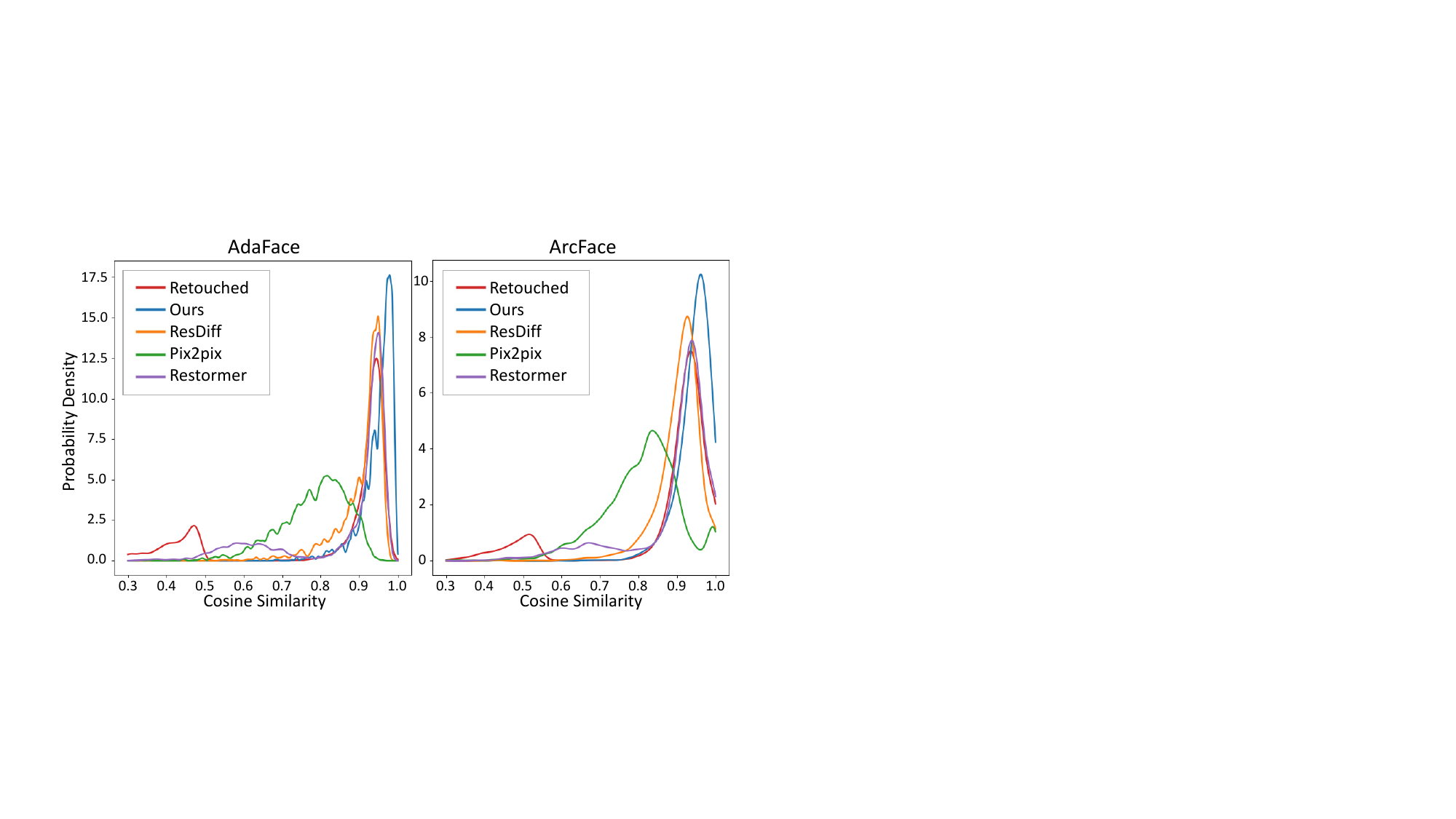}
    \caption{Probability density distribution of facial feature similarity among retouched/restored and original face images.}
    \label{fig:face_rec}
\end{figure}

\begin{table*}[ht]
\centering
\footnotesize
\setlength{\tabcolsep}{5pt} 
\renewcommand{\arraystretch}{1} 
\begin{tabular}{c|cc|cc|cc|cc|cc} 
    \toprule
    \multirow{2}{*}{Methods} & \multicolumn{2}{c|}{White} & \multicolumn{2}{c|}{Smooth} & \multicolumn{2}{c|}{Face} & \multicolumn{2}{c|}{Eye} & \multicolumn{2}{c}{Multi} \\ 
    &PSNR$\uparrow$&SSIM$\uparrow$ & PSNR$\uparrow$ & SSIM$\uparrow$ & PSNR$\uparrow$ & SSIM$\uparrow$ & PSNR$\uparrow$ & SSIM$\uparrow$ & PSNR$\uparrow$ & SSIM$\uparrow$\\ 
    \midrule
    Input & 29.14 & 0.940 & 35.59 & 0.934 & 29.55 & 0.903 & 35.82 & 0.952 & 28.03 & 0.887\\ 
     \midrule
     Pix2pix (CVPR17) & 27.72 & 0.883 & 28.55 & 0.858 & 27.34 & 0.856 & 28.41 & 0.870  & 28.73	& 0.897\\ 
    PCGAN (CVPR18) & 22.47 & 0.684 & 23.53 & 0.639 & 23.04 & 0.627 & 23.63 & 0.634 & 22.17 & 0.581\\ 
    Restormer (CVPR22) & 29.89 & 0.918 & 37.59 & 0.943 & 29.83 & 0.903 & 33.75 & 0.942 & 28.19 & 0.888\\
    DR2 (CVPR23) & 24.33 & 0.723 & 25.97 & 0.719 & 24.39 & 0.712 & 25.71 & 0.719 & 23.87 & 0.711\\
    ResDiff (AAAI24) &26.21 & 0.891 & 35.94 & 0.935 & 29.86 & 0.897 & 35.84 & 0.937 & 28.98 & 0.889\\ 
    \midrule
    MoFRR & \textbf{33.11} & \textbf{0.949} & \textbf{38.06} & \textbf{0.943} & \textbf{31.26} & \textbf{0.913} & \textbf{38.05} & \textbf{0.958} & \textbf{34.47} & \textbf{0.959} \\ 
    \bottomrule
\end{tabular}
\caption{Quantitative comparison of face retouching restoration. All compared models are retrained on our RetouchingFFHQ++ dataset for fair comparison. For each column, all the models are trained and tested on the same single-operated images, or all images for ``Multi''.}
\label{tab:baseline}
\end{table*}

\begin{table}[tb]
\centering
\footnotesize
\renewcommand{\arraystretch}{1}
\begin{tabular}{c|cc|cc} 
    \toprule
    \multirow{2}{*}{Methods} & \multicolumn{2}{c|}{Single} & \multicolumn{2}{c}{Multi}\\
      & PSNR$\uparrow$ &SSIM$\uparrow$ & PSNR$\uparrow$ &SSIM$\uparrow$\\ 
    \midrule
    Input & 34.45 & 0.956 & 26.41 & 0.879 \\
    \midrule
    Pix2pix& 28.18 & 0.881 & 26.68 & 0.886 \\
    PCGAN & 23.18 & 0.626 & 21.30 & 0.569\\
    Restormer& 34.83 & 0.954 & 26.50 & 0.882 \\
    DR2 & 24.72 & 0.709 & 22.63 & 0.676\\ 
    ResDiff & 34.93 & 0.961 & 26.79 & 0.885 \\
    \midrule
    MoFRR &\textbf{36.65} & \textbf{0.971} & \textbf{31.28} & \textbf{0.938}\\
    \bottomrule
\end{tabular}
\caption{Cross-API performance. Models trained on all other datasets are tested on the Portrait subset to evaluate their generalization ability. } 
\label{tab:Cross-API} 
\end{table}

\begin{table}[tb]
\centering
\footnotesize
\renewcommand{\arraystretch}{1}
\begin{tabular}{c|cc|cc} 
    \toprule
    \multirow{2}{*}{Methods} & \multicolumn{2}{c|}{Single} & \multicolumn{2}{c}{Multi}\\
      & PSNR$\uparrow$ &SSIM$\uparrow$ & PSNR$\uparrow$ &SSIM$\uparrow$\\ 
    \midrule
    Input & 32.73 & 0.934 & 28.03 & 0.887\\
    \midrule
     WaveFRR & 35.18 & 0.939 & 31.09 & 0.904\\
     w/o IDEM & 34.34 & 0.936 & 32.94 & 0.915\\
     w/o Degree & 34.43 & 0.937 & 32.95 & 0.921\\ 
     w/o HFCAM & 34.69 & 0.937 & 33.04 & 0.926\\
     w/o shared expert & 35.37 & 0.941 & 32.42 & 0.913\\
     w/o router & 35.63 & 0.942 & 34.19 & 0.947\\
    \midrule
     MoFRR & \textbf{36.12} & \textbf{0.943} & \textbf{34.47} & \textbf{0.959} \\
    \bottomrule
\end{tabular}
\caption{Ablation study over different model modules w.r.t. the ground truth. We fine-tune the models in partial settings till convergence. Test: intra-API.}
\label{tab:ablation}
\end{table}

In \cref{fig:comparison}, we provide a visual comparison with other methods for FRR on the RetouchingFFHQ++ dataset. 
The images are randomly chosen from the subset that contains face images retouched by multiple operations. 
As can be seen in the figures, methods like Pix2pix~\cite{isola2017pix2pix} and ResDiff~\cite{shang2024resdiff} can recover the facial structure to some extent, but fail to generate high-quality face images without artifacts. 
Restormer~\cite{zamir2022restormer} and PCGAN~\cite{chang2018pairedcyclegan} perform better in restoring facial skin color, but struggle with structural restoration.
In contrast, it can be observed that our MoFRR restores the original images more faithfully with fewer artifacts, particularly around the face and eyes. 
Besides, we also interestingly find that the baseline methods that apply the diffusion network generally provides better results compared to the non-diffusion ones. The reason might be that diffusion models also accept the ``divide and conquer" methodlogy in nature, and therefore the image qualities are generally better, as aligned with many recent findings. However, the fidelity of their results are noticeably worse compared to ours, especially from the view that ResDiff in many cases would apply wrong type of retouching.

\Cref{tab:baseline} shows the quantitative comparison of FRR w.r.t. overall PSNR and SSIM.
To better investigate the model performance against different retouching types as well as combined retouching, in the leading four groups we report the performance on images that only has the specific retouching, and in the last group (named ``Multi"), the performance on images with two or more retouching operations.
From the results, our method outperforms other methods on the four single-operated subdatasets. 
Although PCGAN~\cite{chang2018pairedcyclegan} exhibits excellent performance on the makeup transfer task, it struggles on the novel task of FRR with PSNR of around 23 dB. DR2~\cite{wang2023dr2} performs well on face restoration but struggles with FRR due to the addition of excessive detail noise, resulting in higher image quality but lower fidelity.
Other baselines have their own advantages in different subsets but do not perform as well and as comprehensively as MoFRR, particularly on the whitening subset where MoFRR achieves 3.97 dB improvement in PSNR. 
Particularly, our proposed MoFRR demonstrates superior performance compared to all baselines on the multi-operated subset. The distortion effects of multi-operated images are more pronounced. MoFRR effectively restores retouched face images, which surpasses the second-best method by 5.49 dB, achieving the highest improvement of 6.44 dB in PSNR. Multi-operated images better represent actual application scenarios, indicating that our method has significant advantages in practical use. The significant improvement observed in the multi-operated subset further validates effectiveness of our MoE framework.

\Cref{fig:single} provides the visual results generated by specific experts. 
To facilitate a clearer visual observation, we provide a sub-figure in each row as a zoom-in on some region-of-interest areas.
It can be observed that four operation-specific experts perform well in restoring the targeted retouching operation, 
where we indeed see that the correspoding retouching is successfully reverted while other types of retouching operations that the  expert is not specialized in are kept unchanged.
We also see that the shared expert is able to do multiple types of retouching removal in parallel, but the generated result is not close to the ground-truth images.
This finding is generally in line with our hypothesis that though one expert is not enough yet could steadily contribute to the overall quality improvement via seeing more retouched samples compared to the specialized experts. 

\noindent\textbf{Impact on Face Recognition.} 
\Cref{fig:face_rec} illustrates the probability density distributions of facial feature similarity between retouched/restored and original images. Each curve represents the cosine similarity distribution of a specific method’s results compared to the original faces, where a value closer to 1.0 indicates near-perfect feature alignment.
The blue curve of MoFRR exhibits a sharp peak near 1.0 across both AdaFace~\cite{kim2022adaface} and ArcFace~\cite{deng2019arcface}, which means the faces recovered by using our proposed method are more similar to the original faces at the feature level. The narrow width of the blue curve’s peak further highlights its high confidence in restoring identity features.
While existing methods ResDiff~\cite{shang2024resdiff} and Restormer~\cite{zamir2022restormer} improve over retouched images, their curves show flatter distribution and lower peak values than ours.
The curve of Pix2pix~\cite{isola2017pix2pix} has the lowest peak value and the flattest distribution, implying feature misalignment which may be caused by the heavy artifacts.
Besides, the long left tail of the red curve indicates some retouched images are extremely processed with severe identity degradation. All methods can restore the extreme cases to some extent, while our MoFFR suppress the left-tail density more effectively.


\begin{figure}[!t]
    \centering
    \includegraphics[width=1\linewidth]{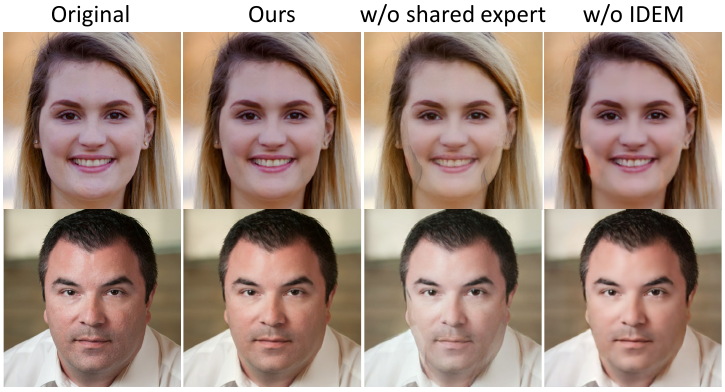} 
    \caption{Qualitative comparison of the ablation studies, which proves the necessity of employing shared expert like DeepSeek and the IDEM design.}
    \label{fig:ablation}
\end{figure}

\subsection{Cross-API Performance} 
In a real-world scenario, training data and testing data are likely to come from different sources, so cross-API performance analysis helps us probe into the estimated performance of models in real-world conditions.
From the results shown in \cref{tab:Cross-API}, we observe that while there is a modest degradation in overall PSNR, MoFRR consistently delivers commendable restoration outcomes. 
Compared to the baselines, which achieve poor performance on both single- and multi-operated images, MoFRR demonstrates a notable advantage with improvements of 1.2 dB and 3.87 dB.
This substantial performance gain, particularly for multi-operation images, can be attributed to the specialized expertise of multiple expert networks within MoFRR. 

\subsection{Ablation Study}


\textbf{\noindent\textbf{Architectural analysis.}}
The results are shown in \cref{tab:ablation}. 
It can be observed that removing any one of modules in WaveFRR degrades performance compared to the full model. Both the Degree Estimator and IDEM handle distortion predictions, with the former focusing on global predictions and the latter on pixel-level predictions. The removal of IDEM results in significant performance degradation due to its refined iterative predictions. The combination of the two modules makes the prediction more effective by indicating the location and extent of retouching, as well as HFCAM demonstrates unique effectiveness in artifact removal.
The performance degradation of removing the shared expert mainly occurs in multi-operated cases due to its effect on global structural restoration.
In contrast, removing the router has minimal impact on performance, as the Degree Estimator can reduce the impact of false activation while the router contributes positively to enhancing efficiency.

\modify{\noindent\textbf{Comparison with Single-Expert Mode.}}
We also test the performance where we use one single WaveFRR as the whole MoFRR.
Results show that it performs well on single-operated images, but limited when restoring all operations simultaneously.
The multi-expert mechanism also offers flexibility that once novel types of face retouching appear, a plug-in of corresponding expert networks would mitigate the huge cost of fully retraining the model.

\section{Conclusion}
\label{sec:conclusion}

We propose face retouching restoration, a novel task of recovering the original faces from retouched ones. We propose a Mixture of Diffusion Models for FRR, including specialized WaveFRR models that uses a dual-branch structure, i.e., a low-frequency recovery using DDIM, and a high-frequency refinement guided by the restored lower subbands. 
Extensive experiments on our dedicated RetouchingFFHQ++ dataset prove the effectiveness of our model on FRR against varied combination of retouching operations.

\noindent\textbf{Acknowledgment}~This work was supported by Ant Group through CCF-Ant Research Fund.

{
    \small
    \bibliographystyle{ieeenat_fullname}
    \bibliography{main}
}

\end{document}